# The Food Recognition Benchmark: Using Deep Learning to Recognize Food on Images

Sharada Prasanna Mohanty[a,1], Gaurav Singhal[b], Eric Antoine Scuccimarra[c], Djilani Kebaili[d], Harris Héritier[d], Victor Boulanger[d], and Marcel Salathé[a, d]

[a]Alcrowd Research, Alcrowd; [b]Otto von Guericke University Magdeburg; [c]Société des Produits Nestlé, Nestlé Research Center; [d]Digital Epidemiology Lab, EPFL (École polytechnique fédérale de Lausanne)

**The automatic recognition of food on images has numerous interesting applications, including nutritional tracking in medical cohorts. The problem has received significant research attention, but an ongoing public benchmark to develop open and reproducible algorithms has been missing. Here, we report on the setup of such a benchmark using publicly available food images sourced through the mobile MyFoodRepo app. Through four rounds, the benchmark released the MyFoodRepo-273 dataset constituting 24,119 images and a total of 39,325 segmented polygons categorized in 273 different classes. Models were evaluated on private tests sets from the same platform with 5,000 images and 7,865 annotations in the final round. Top-performing models on the 273 food categories reached a mean average precision of 0.568 (round 4) and a mean average recall of 0.885 (round 3). We present experimental validation of round 4 results, and discuss implications of the benchmark setup designed to increase the size and diversity of the dataset for future rounds.**

Food Recognition | Deep Learning | Instance Segmentation | Crowdsourcing

For almost all of human history, the main concern about food centered around one single goal: to get enough of it. Only in the past few decades has food ceased to be a limited resource for many. Today, food is abundant for most - but not all - inhabitants of high- and middle-income countries, and its role has correspondingly changed. Whereas the main goal of food used to be to provide sufficient energy, today, the main public health challenges are the avoidance of excessive calories, and the nutritional composition of diets.

The health burden of diets at the population level is increasingly well understood. Diets leading to excess weight and obesity are thought to be at least partially responsible for chronic disease mortality and morbidity associated with non-communicable diseases (NCD) (1). The nutritional composition of diets is strongly linked to health outcomes. For example, diets high in sodium, low in whole grains, low in fruits, low in nuts and seeds, and low in vegetables, are associated with the highest number of deaths attributable to diet at the global level. As the "EAT-Lancet Commission on Health Diets from Sustainable Food" noted, unhealthy diets now pose a greater risk to morbidity and mortality than unsafe sex, and alcohol, drug, and tobacco use combined (2).

While the link between diet and health at the population level is increasingly clear, there is at the same time a growing understanding that things are not quite so simple at the individual level. How exactly individuals' diets affect their health is only poorly understood. For example, recent research focusing on a post-meal glucose, a specific biomarker and risk factor of type II diabetes, has shown that there is substantial individual variability in glycemic response to identical meals (3). This suggests that the effect of diet on health outcomes is modulated through various other factors (such as the microbiome composition), and that generic diet recommendations may be of limited use. As a consequence, the concept of personalized nutrition has emerged, which aims to adjust diets to the individual in order to maximize the positive effect on health outcomes.

Furthermore, the increasingly recognized importance of the microbiome and its relation to diet have put nutrition once again on the radar of cutting-edge medical research. For example, it's been shown that the gut microbiota affects the immune system, drug metabolism, and the effect of immune therapies against cancer (4–6). The fact that diet affects the gut microbiome is well established (7), but the exact mechanisms are not yet well understood. Recent years have seen numerous studies addressing the causal relationships between food and health outcomes, and more research is expected in this area (8).

However, research in dietary patterns faces an important obstacle: the accurate measurement of food intake. Traditionally, the measurement of food intake has involved methods such as food frequency questionnaires, or 24-hour recalls. These methods have been widely used, but are known to have substantial problems. In recent years, digital alternatives have appeared, hoping to leverage the ease-of-use of many mobile applications. Applications that simply allow study participants to enter their food intake manually via text, however, do not provide significant advantages in terms of ease-of-use over paper-based methods, as text entry on mobile devices is cumbersome. More promising are applications that scan the bar codes of food products, and subsequently extract the nutritional content from associated databases. Perhaps the most promising are applications leveraging the enormous advances in image recognition in the past few years. The act of taking a picture of food presents the least burden for a participant, but it provides a formidable technical challenge to correctly extract the nutritional content from an image alone, as it requires the recognition of the food and the amount from image data.

Here, we provide an overview of an approach developed to tackle this problem. The approach is based on the notion that accurate food image recognition is feasible, but can't easily be solved in one single go, and instead requires iterative improvements over time. It is further based on the notion that a crowd-sourced approach with properly aligned incentives can efficiently leverage machine learning know-how around the world, providing a much broader intellectual attention to the problem than the classical "single group" research approach. Last, but certainly not least, the approach is based on open data and open source models, where the necessary data to train the models are provided as open data (licensed under the Creative Commons CC-BY-4.0 license), and the code of the submitted models have to be released under an open-source license of choice in order to be eligible for prizes. Currently, the images in the open data set have been collected through the use of the myFoodRepo mobile app, which is used in medical cohorts in Switzerland, and increasingly in other countries as well.

The result of this approach is the continuous AIcrowd Food Recognition Benchmark, which we describe here in detail. We then describe solutions to the problem provided by the winning submissions to the recently finished round 4 of the benchmark. Finally, we will describe open challenges, and next steps.

## 1. Related Work

The use of machine learning for recognition of food in images has had a renewed momentum because of introduction of novel datasets like UEC-FOOD101 (9), UEC-FOOD256(10), UEC-FoodPix (11), UEC-FoodPixComplete (12), etc. The UEC-FOOD101 and UEC-FOOD256 datasets primarily focus on the food classification task with the assumption that each image has a single food item, and the goal is to be able to predict a food class for each image received. Both datasets also include bounding box locations for the food item in the picture to enable researchers attempting object detection. The UEC-FoodPix and UEC-FoodPixComplete datasets introduces the Food Image Segmentation task with a dataset of 10,000 images which include multiple food items per image, and associated pixel wise segmentation masks. The Food Image Segmentation task involves being able to accurately identify multiple food items in the same image, and draw accurate boundaries for each of the food items. With the maturity of Deep Learning, the Food Image Segmentation task has started to deliver promising results on a task that was previously considered extremely complex. Pishva et al. (13) demonstrate the feasibility of segmenting and classifying 73 kinds of bread from images of break on a plate. While the results are impressive, the dataset is collected using a fixed camera setting, where the bread is put at the center of a unicolor plate. With uniform background, the segmentation task is a relatively easier problem that the real world counterparts. A usable Food Image Segmentation approach will have to address numerous real world aspects of the problem, including, but not limited to, multiple food items on a place, different shapes/textures of the food items, overlapping food items, images obtained from arbitrary camera placements, and uneven lighting conditions. Ciocca et. al. (14) introduce the UNIMID2016 dataset which includes the multi-food setting, however the images were taken in a laboratory setting where each food item is placed on a separate plate and all the plates are placed on a tray - making the segmentation task relatively easier. (15) introduced the usage of deep learning in food image recognition, by applying YOLOv2, DarkNet-19 on the UNIMIB2016 dataset to eventually obtain a precision of 0.841. With the emergence of large datasets, and cheap compute, deep learning algorithms have become a popular choice for problems where the principal modality of the data are images or text. Mask R-CNN (16) has been a popular approach for instance segmentation tasks. Ye et. al. (17) used Mask R-CNN with MobileNet and ResNet for food segmentation by handpicking 10 food categories from the MS COCO dataset. Using the said dataset, Ye. et. al. (17) did a comparative analysis of deep learning architectures (Mask R-CNN with MobileNet and ResNet), vs Multi-SVM; eventually demonstrating that Mask R-CNN with ResNet outperformend Multi-SVM by a significant margin. Freitas et. al. (18) reported results from various experiments with different deep learning architectures (including Mask R-CNN, DeepLab V3, SegNet, ENet) on a proprietary dataset of Brazilian food items. They demonstrate that Mask R-CNN outperforms the rest of the approaches, with a mAP of 0.87, whereas the rest of the methods scored a mAP of less than 0.79.

The performance of modern deep learning approaches highly depends on the size and the quality of the dataset. The results of the instance segmentation task in the MS COCO Benchmark (19) only reiterate the same thought. Large real world datasets of food images along with high quality human annotations are finally enabling classification, detection and instance segmentation of food items as accessible problems which yield usable tools.

## Dataset

**Source Data.** The data used in this study was made available by the myFoodRepo app users between **July 7th 2018** and **June 8th 2020**. The dataset consists of **24,119 images** containing a total of **39,325 segmented polygons**. The food images are categorized in **273 different classes** with **at least 35 annotations per class**. The pictures taken via the myFoodRepo app are private by default, but users can choose to make their anonymized images public for research purposes. Since late 2018, the number of public pictures has been steadily growing because the myFoodRepo app is used by the participants of the Swiss Food and You cohort, a personalized nutrition study focusing on postprandial glycemic responses. The share of public/private images has been growing in similar magnitude, shown in Figure 2.

The myFoodRepo app offers three ways to track food intake:

- Manual entry
- Barcode scan
- Image of food taken with camera

The last case - data input in the form of images taken by the phone's camera - represents the majority of data input (~90%) through the myFoodRepo app.

**Instance segmentation and class labelling.** The resulting images are initially analysed by an algorithm that performs instance segmentation and food class prediction of the segmented items as shown in Figure 1. The segmentations and food classes are further manually assessed by human annotators via the myFoodRepo annotation web interface. The human assessment mainly consists of redrawing instance segment annotations and correcting class assignments. If the algorithmic analysis led to an incorrect or missing segmentation, or predicted the wrong food class, the human annotator will provide manual corrections. This human correction and verification step is critical for quality assurance. The myFoodRepo app is used in multiple medical cohort studies, and a high accuracy of the data annotation is of paramount importance. At the same time, every additional verified annotation helps grow the high-quality training dataset which is subsequently used to improve the deep learning models for food image recognition. These improved models can then be used by the myFoodRepo app to better assist the annotators. Over time, this cycle is expected to decrease the time spent by human annotators per image, and thus allows for a higher and higher data throughput while maintaining quality.

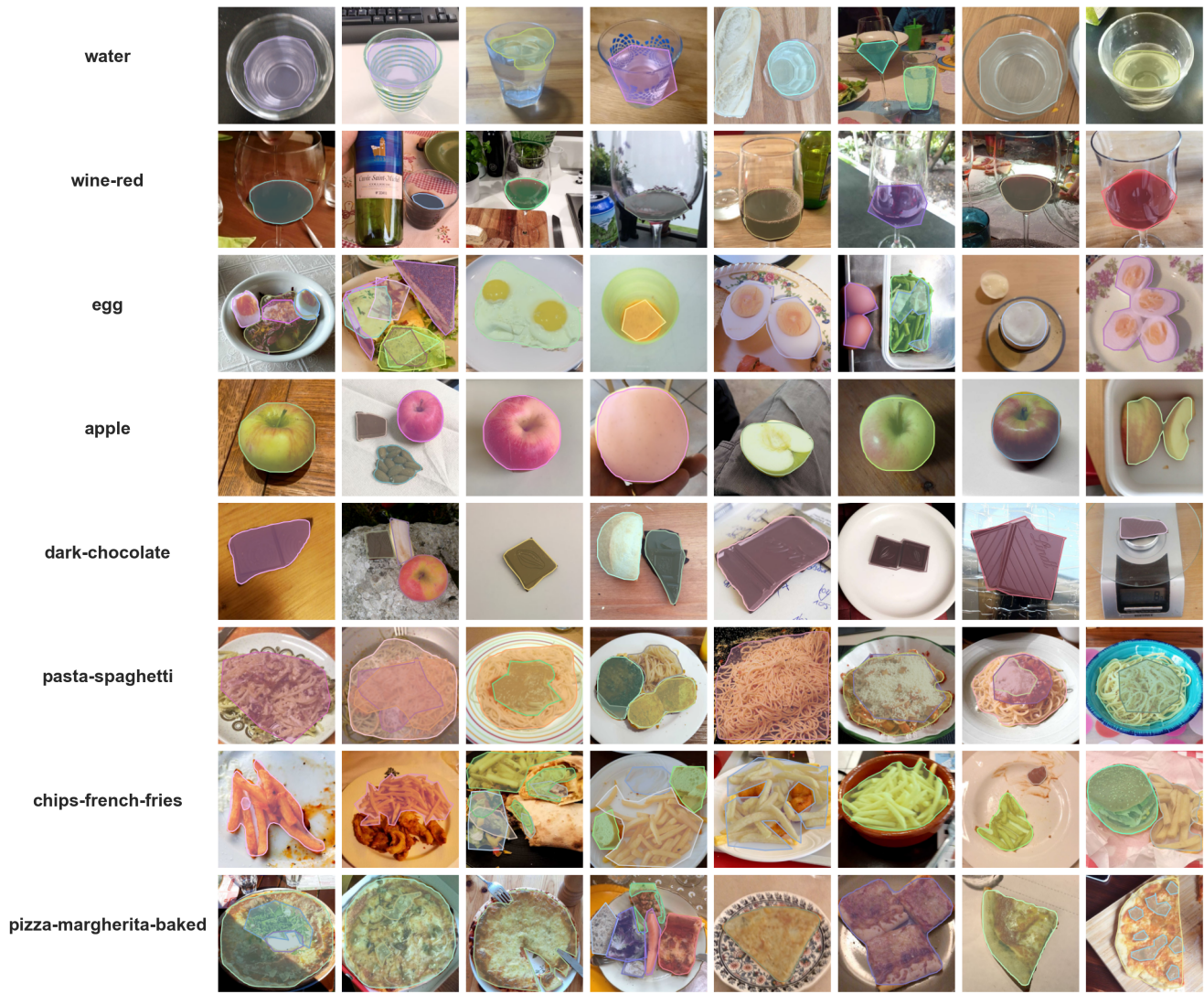

**Fig. 1.** Sample images and corresponding annotations for some of common food classes in myFoodRepo dataset.

This assisted annotation approach leads to a bi-modal distribution in terms of the number of points per annotated polygon across all the annotated polygons as shown in Figure E.1. Human annotators often draw polygons consisting of around 8 points for reasons of efficiency, whereas the instance segmentation algorithms predict polygons with much a higher number of points.

The food classes were primarily derived from the first Swiss national survey on nutrition, "menuCH" (20) that was conducted using the validated Globodiet tool. However, during the productive use of the myFoodRepo in medical cohorts in Switzerland, the scope of this food class list was found to be insufficient, and new food classes where created on the go when myFoodRepo users uploaded pictures of food items not belonging to any yet existing class.

The human annotation was performed by expert annotators in the Digital Epidemiology Lab specifically trained for this task. In case of uncertainty, an annotator has the option to communicate with the myFoodRepo user through the app to ensure the annotation is accurate. The combination of an initial annotation by the algorithm, the annotators' expertise, the possibility to interact with the user for further clarifications, and the need for efficiency, each image was only handled by one human annotator.

**A. Benchmark.** The food recognition benchmark was designed as an initiative to engage a broader community of researchers to train better models for food recognition, which in turn are used to assist the annotators, catalysing the process of creating much larger annotated datasets needed to train better models for food recognition.

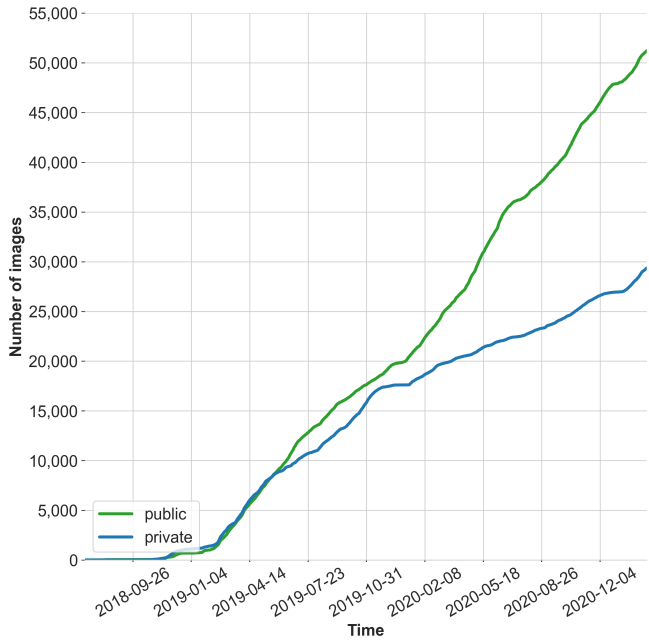

**Fig. 2.** Cumulative number of images collected via myFoodRepo App (including classes not considered in this paper)

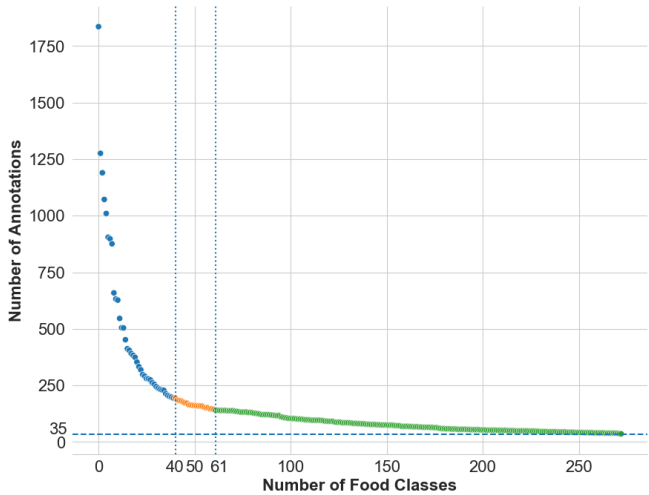

**Fig. 3.** Distribution of number of annotations available across the food classes in the dataset. The individual food class names are not mentioned for readability. The public release of the dataset was done in 3 phases - the first phase of the public release consisted of the images from the top-40 classes, the second phase of the public release consisted of the images from the top-61 classes, and the third phase of the public release consisted of the images from the top-273 classes.

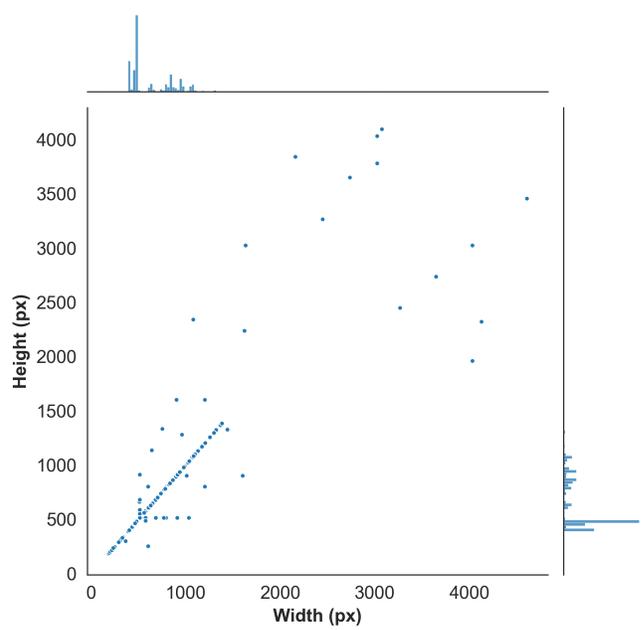

**Fig. 4.** Distribution of the image width and image height in the public dataset.

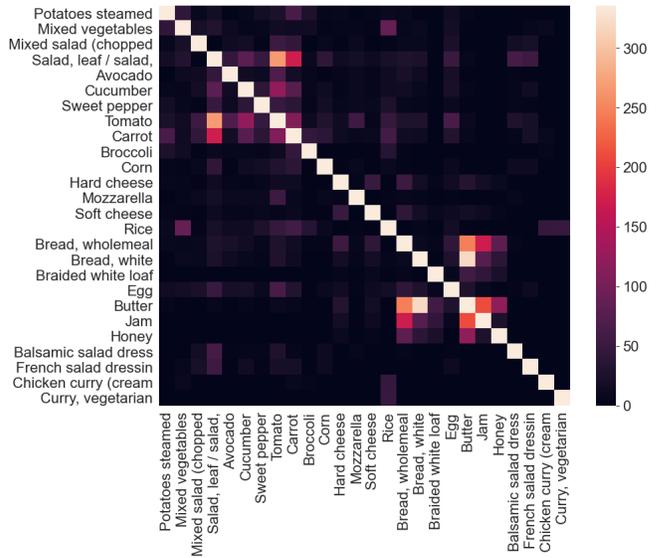

**Fig. 5.** Co-occurence matrix of 26 food classes which have a co-occurrence of more than 40 instances across the publicly available dataset. When computing the co-occurrence matrix, the self co-occurrence values were not considered, and for illustration purposes, it has been artificially set to a value slightly above the maximum co-occurrence values across all the classes in consideration.

The benchmark relies on the myFoodRepo app's users to report their daily food intake (Figure 6-1). The images collected by the myFoodRepo app are then processed by the Food Recognition API (Figure 6-2) to generate instance segmentation annotations for the images (and classification ???). The annotated images are then passed on to a team of annotators (Figure 6-3) who enrich the quality of those annotations by redrawing certain segments, correcting mislabelled food classes, or manually annotating instances of food items that were missed by the algorithm in the Food Recognition API. This process generates the myFoodRepo dataset. The my-FoodRepo dataset has a public component (Figure 6-4b) and a private component (Figure 6-4a), depending on the privacy preferences of the users of the app. The public dataset, along with the corresponding annotations, is provided as a training set for the community (Figure 6-5) of participants, who analyze the dataset and train their models (Figure 6-6) using the dataset. The trained models are then subsequently submitted to AIcrowd for evaluation (Figure 6-7), which generates the leaderboard for the participants. The leaderboard (Figure 6-8) acts as a feedback for participants who use the feedback signal to improve their models, and who continue to submit their improved models to the AIcrowd evaluators. The AIcrowd evaluators use the private version of the myFoodRepo dataset (Figure6-3a) to evaluate the submissions, hence ensuring that the submitted models are evaluated in a setting that's a close to the real world as possible. The best models on the leaderboard are automatically deployed as an API (Figure6-2) by the AIcrowd evaluators. The API then continues to assist the annotators to annotate a larger number of images, more accurately, more efficiently and much faster, leading eventually to a much larger public training dataset available for participants to train their models. The improved models are then submitted by the participants to the evaluators, which feed an improved leaderboard and an improved food recognition API for the annotators.

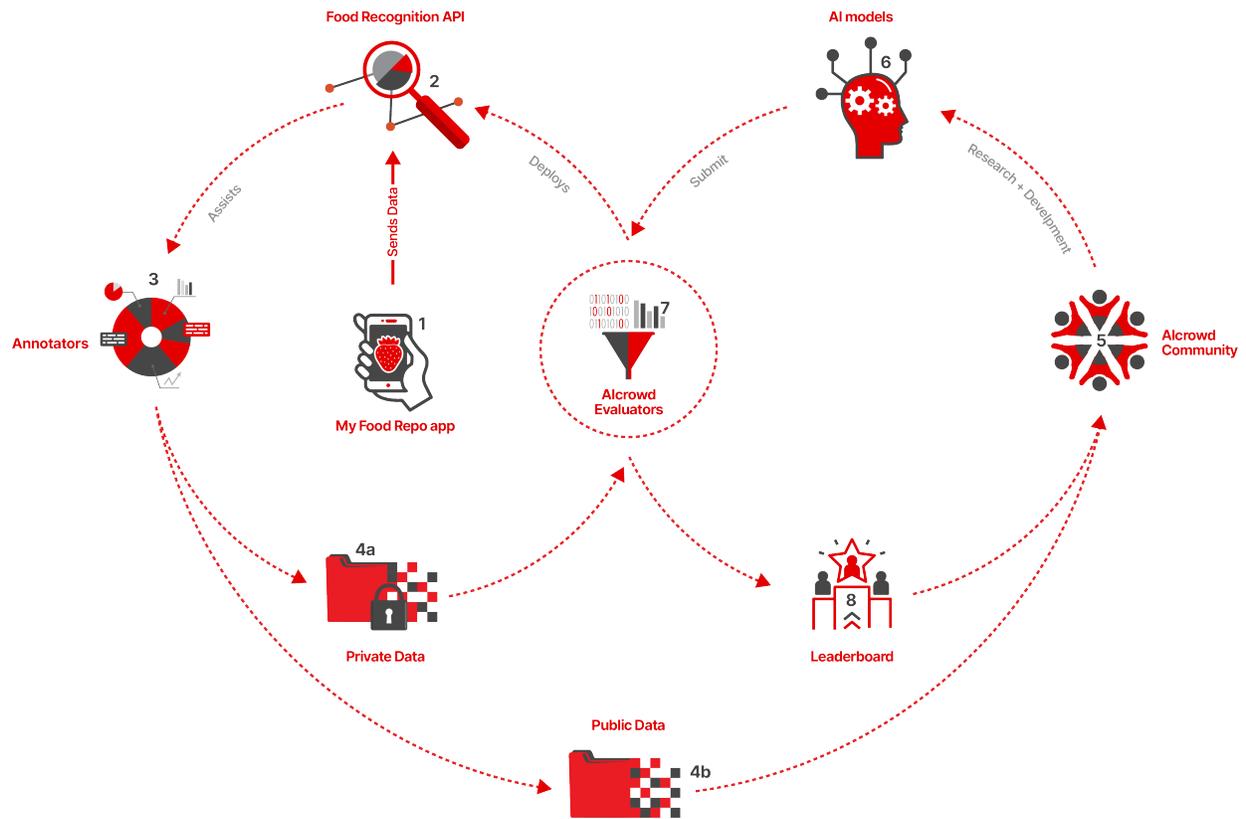

**Fig. 6.** A high level design for Food Recognition Benchmark. Section A contains a detailed explanation of the all the individual components of the Benchmark and how they interact with each other.

The first iteration of the Food Recognition Benchmark was organized across 4 different rounds :

**Round 1.** Round 1 of the Food Recognition Benchmark started on 10th October 2019 and ended on 31st December 2019. This round focused on the 40 food categories which had at-least 35 annotations each. The training dataset(`train-v0.1`) consisted of 5,545 images and 7,735 annotations. The test dataset(`test-v0.1`) had 1,959 images and 2,176 annotations.

**Round 2.** Round 2 of the Food Recognition Benchmark started on 28th January 2020 and ended on 17th May 2020. This round focused on the 61 food categories which had at-least 35 annotations each. The training dataset(`train-v0.2`) consisted of 7,949 images and 11,468 annotations. The test dataset(`test-v0.2`) had 3,115 images and 3,667 annotations.

**Round 3.** Round 3 of the Food Recognition Benchmark started on 1st September 2020 and ended on 8th January 2021. This round focused on the 273 food categories which had at-least 35 annotations each. The training dataset(`train-v0.4`) consisted of 24,119 images and 39,325 annotations. The test dataset(`test-v0.4`) had 5,000 images and 8,061 annotations.

**Round 4.** Round 4 of the Food Recognition Benchmark started on 15th January 2021 and ended on 1st March 2021. This round continued to focus on the same problem formulation as Round-3 and had 273 food categories which had at-least 35 an-notations each. The training set was the same as Round-3, and a new test set containing 5,000 images and 7865 annotations was introduced.

The training dataset released as a part of Round 3 and Round 4 constitute the **MyFoodRepo-273** dataset and is accessible at : https://www.aicrowd.com/challenges/food-recognition-challenge/dataset_files.

The test dataset across all the rounds will not be publicly released as it is composed of the private myFoodRepo dataset. The benchmark will continue to use the round specific test sets to evaluate future submissions to the specific rounds.

**B. Evaluation Metrics.** We build up on the tradition established by the PASCAL VOC Challenge (21) to continue using **mean average precision (mAP)** and **mean average recall (mAR)** as the metrics to evaluate the results. In both the cases, we use an Intersection over Union (IoU) $> 0.5$ as the qualifying criteria for computing the precision and recall. The IoU in PASCAL VOC is computed on bounding box predictions for object detection tasks, and in case of the Food Recognition Benchmark, we compute the IoU from the overlap of the instance segmentations instead.

**C. Benchmark Statistics.** The first iteration of the Food Recognition benchmark saw participation from **1,065 participants** (as of 17th April, 2021) from **71 countries**. A total of **2,603 submissions** were made, amounting of approximately **2.5 TB** of user-submitted code and models. The best model in round 1 had a mean average precision of **0.573** and a mean average recall of 0.831. The best model in round 2 had a mean average precision of **0.634** and a mean average recall of 0.886. The best model in round 3 had a mean average precision of **0.551** and a mean average recall of 0.885. The best model in round 4 had a mean average precision of **0.568** and a mean average recall of 0.767.

**D. Methods.** The methods section summarizes the best solutions that were explored in the context of the Food Recognition benchmark. Most of the instance segmentation methods follow the same anatomic pipeline and usually consist of two stages. The first stage of the pipeline focuses on extracting the feature maps from the input image and later use these features to propose the interesting regions that may contain the object. The second stage is a parallel network of different predicting heads, where Classification, Bounding Box, Masking are done for the interesting region. The participants in the competition explored various architectures including Mask R-CNN (16), Hybrid Task Cascade (HTC) (22), Cascade R-CNN (23), DetectoRS (24).

In the experiments reported in this manuscript, we evaluate a subset of the said architectures : Hybrid Task Cascade (22), Mask R-CNN(16).

We observe that the cascade models provided significantly better performance than the other models, with HTC providing the best performance. This led us to focus on using HTC with two different backbones - ResNet 50, ResNet 101 (25). We also provide comparative study of the performance of Mask RCNN model with three different backbones - ResNet 50, ResNet 101, and ResNext 101 (25).

Mask R-CNN is a simple and flexible extension of Faster R-CNN (26) to perform instance segmentation. Faster R-CNN can efficiently detect objects in an image, and with a few changes, Mask R-CNN can do the same and also able to generate a good segmentation mask for each instance of an object. Mask R-CNN does this by adding a mask head in parallel to the head responsible for making classification and regression. It is among the first few methods to perform end to end instance segmentation end-to-end. Stage 1 of Mask R-CNN extracts the feature maps using the help of an FPN (Feature Pyramid Network) backbone architecture with the help of well-known CNN-based architectures such as ResNet50, ResNet101, etc. FPN network creates a lateral connection with the various residual blocks. It provides a top-down pathway by using semantically rich layers to make higher-resolution layers. The RPN network then proposes thousands of regions which are then checked if they are a foreground or background. Regions predicted as background are discarded and foreground regions are called as Regions of Interest (RoI). These RoIs are then passed through the RoIAlign (16) which aligns the extracted features with the input. The RoI is then passed to the second stage which performs the classification and predicts a bounding box of the proposed region. The same RoI is passed to the masking head in parallel which predicts the binary mask for the object in RoI. Mask R-CNN is a single-stage detector, trained with a lower threshold such as 0.5 IoU. It results in producing many noisy predictions.

(23) proposed the very first multi-stage object detection architecture which encapsulates a sequence of detectors. They are trained with growing IoU thresholds to enables the network to make smart decisions over close false positives. The sequence of detectors also called a cascade is proved to work better than all other single-stage object detectors. Even after this intuitive idea of the cascade, it misses the relationship between detection and segmentation. (22) on the other hand take full advantage of the reciprocal relationship between the different heads, here, detection and segmentation. Hybrid Task Cascade is different in several aspects. Instead of executing bounding box regression and masking in parallel, it interleaves. It includes a straight path that reinforces the flow of information between mask branches by feeding the previous stage's mask features to the present one. It also adds semantic segmentation and combines it with the bounding box and masking branch. It seeks to investigate more contextual information. Overall, these changes to the framework architecture increase the flow of information not only between stages but also across tasks.

FPN uses the lateral connections to the bottom-up layers in CNN architecture which helps in looking at the image once or twice. DetectoRS (24) is a state-of-the-art instance segmentation algorithm, which is an extension of (22) and focuses only on the backbone architecture of Hybrid Task Cascade. At the macro level, they proposed RFP (Recursive Feature Pyramid) which is built on top of FPN (Feature Pyramid Network). RFP creates feedback connections from the FPN layers into the bottom-up backbone layers. Unrolling this recursive network to a sequential will help look at the input image more than twice. This change recursively improves the FPN to generate a more powerful feature representation, improving the performance. At the micro-level, they propose Switchable Atrous Convolution (SAC). It uses the switch functions to gather the results obtained by convolving the same input features with varying atrous rates.

***D.1. Preprocessing.*** Exploratory data analysis on the annotated images in the dataset revealed some issues. A very small number of images containing disoriented instance segmentation masks were removed from the training set. Numerous annotations included bounding boxes which did not match the corresponding segmentation masks. This would present a problem for cascade models which used both the masks and the bounding boxes as labels, so all of the bounding boxes for all annotations in both the train and validation sets were re-computed to match the segmentation masks.

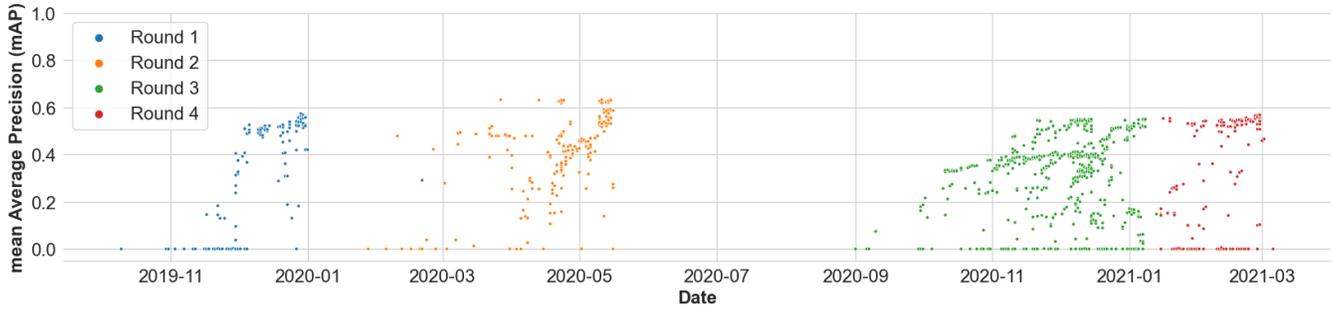

**Fig. 7.** Distribution of the mean Average Precision (mAP) scores from the submitted results across time, and across all the four rounds of the benchmark.

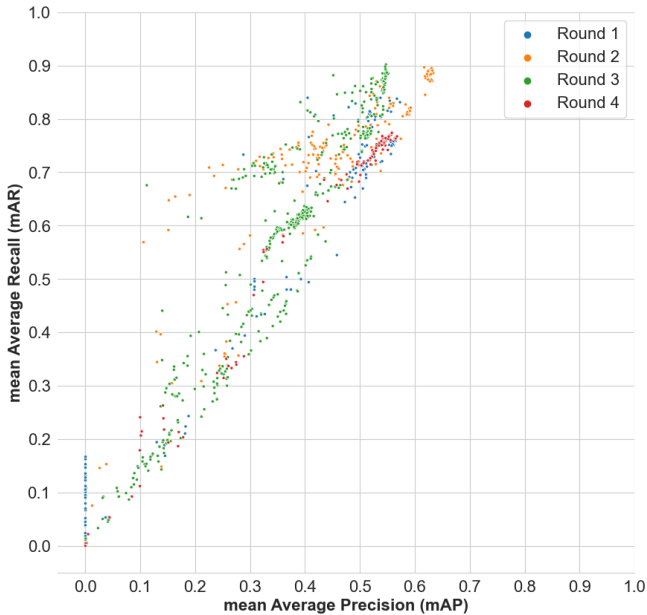

**Fig. 8.** Distribution of the mean Average Precision (mAP) scores and the mean Average Recall (mAR) scores across all the submissions made in the benchmark.

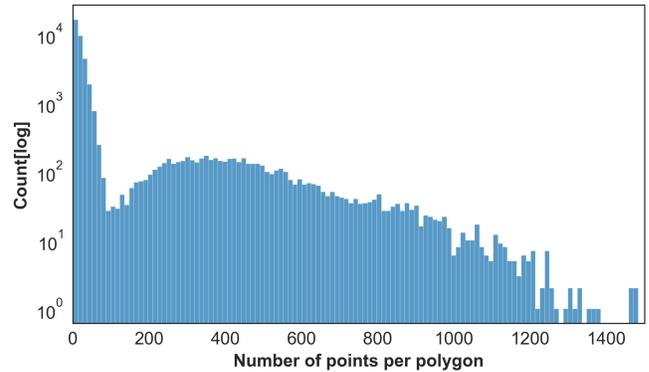

**Fig. 9.** Distribution of the Number of Points across all the polygons(instance segmentation annotations). For the sake of clarity, this plot excludes a small fraction of the annotations with number of points higher than 1500.

***D.2. Data Augmentation.*** To avoid over-fitting, we used fairly aggressive data augmentation, powered by the albumentations library(27). The dataset contains images in a wide range of sizes, ranging from 183x183 to over 4000x4000. However, 50% of the images are between 480x480 and 852x852, with only 20% smaller than 400x400 or larger than 973x973. Accordingly, we used multi-scale training with the sizes ranging from 480x480 to 960x960. Random horizontal flip was used along with random rotation. As some objects in images (such as glasses of water) depend on their orientation, we kept rotation to between 4 and 12 degrees. Random shifting and scaling were also used, as well as perspective augmentations. As the resolution of the training images varies greatly, random JPEG compression was applied. Random hue, saturation, brightness, contrast, blur, median blur, and Gaussian noise were also used.

Many of the classes are very similar and very difficult to distinguish even for a human. On the assumption that the only way to distinguish between such classes as peppermint tea and herbal tea was through subtle differences in coloring, we kept color-related augmentations to a minimum. Accordingly, a small amount of RGB shift was used with a low probability.

***D.3. Experiment Configuration.*** We altered mmdetection's default learning rate decay schedule to use a gamma of 0.5 instead of 0.1, but the learning rate was decayed using a more frequent schedule than the default.

Test time augmentation was performed using mmdetection's MultiScaleFlipAug to evaluate each image at 3 sizes and two horizontal flips. The sizes were chosen as the lower and upper ends of the sizes used for multi-scale training and the midpoint between those two values. This means that each image was evaluated 6 times. This produced a large number of detections, many of them overlapping, resulting in a large number of false positives.

***D.4. Model Ensembles.*** Casado-Garcia et al. (28) propose methods to ensemble the results of multiple object detection models. Their method groups detections by class and overlapping IOU > 0.5, then uses one of several methods to determine which groups to keep, and finally retains the detection with the highest score from each retained group. While we are interested in the masks, their method operates on bounding boxes.

However, converting RLE masks to binary masks and then calculating the IoU of the binary masks required too much time to be usable within the inference time constraints. While not ideal, we found that using the bounding boxes to group and filter the predictions was an acceptable solution, provided that the bounding boxes provided as labels had been corrected to match the masks. This was another advantage of the cascade models which output both masks and bounding boxes.

Our methodology for merging predictions used the same grouping function proposed by Casado-Garcia et. al (28), but the subsequent steps were tailored for this problem. Since many of the masks have unusual, non-compact shapes, which results in large bounding boxes, we wanted to favor larger boxes over smaller ones. To do this, we assigned each detection a weight which was calculated by multiplying the area by the score, then non-maximum suppression (NMS) was performed using this weight rather than the confidence score. After each group had been reduced to one detection, the remaining detection was assigned the maximum of the scores in the group. Our original idea had been to assign the remaining detection the mean of the scores in the group, and while this improved the mAP on the validation data, it reduced the scores on the test data, which implies that some of the correct detections have very low scores, a hypothesis which is supported by comparing detections on the validation set to the ground truth.

Once this method for merging predictions had been verified to improve performance for a single model, we decided to try to apply it to the outputs of multiple models. Casado-Garcia et. al (28) had proposed only retaining detections made by more than one model (or, equivalently, groups with length 1) which seemed like a good approach but produced significant reduction in mAP when applied to this problem. However, retaining groups with only one detection but reducing the score of that detection by half improved the mAP. We believe this is related to the difficulties related to classification, which is reflected by the fact that often, many correct detections have very low confidence scores.

Many experiments were performed using different combinations of architectures, backbones, amounts of augmentation, limits of the multi-scale training sizes, and different weights of the classification loss. From all of these experiments, the models and epochs with the best performance on the validation set were then evaluated as members of a three model ensemble. While ensembles of more models performed better, any more than three models took longer than the allowed evaluation time to complete inference.

***D.5. Negative Results.*** We also explored numerous approaches, which did not yield significant results. We would like to document these negative results for completeness.

We trained a separate image classification model and used its predictions to condition the predictions made by the object detection and instance segmentation predictions. This approach did not lead to any significant gains in the detection results. We also attempted to create an ontology of the food classes, and used the same to hierarchically filter out the outputs. Even if we had strong expectations of result improvements, there were no significant gains observed. We used the co-occurrence matrices of all different food classes to identify similar food categories (i.e. different types of tea, or different types of bread) and then attempted to group together predictions from the same category as the dominant category in the group, or as a randomly selected category in the group. Separately, we also used the co-occurrence matrices to filter out the final predictions. Both the approaches did not yield any significant improvement in the results. We attempted to remove non-overlapping predictions of the same class, tried using the agreement between various models to determine which detections to keep, and to remove or merge detections which are completely contained in a separate detection of the same class - without any significant gains in the detection results in any of the approaches.

We also explored using class weights for the classification loss, which surprisingly did not work. We believe that the main difficulty in applying any sort of post-processing to the results is the fact that the correct classifications often have very low scores. The success in the segmentation task, and the difficulty in the classification task indicates the need for much larger training dataset sizes, especially in case of the class distributions across 273 categories (Not clear???).

## E. Results.

***E.1. Benchmark Results.*** Table 1 outlines the best mAP and mAR scores received from the best submissions across the four rounds of the Food Recognition Benchmark. In round 1, the problem formulation focused on 40 food categories, the best submission received a mAP of 0.573 and a mAR of 0.831. In round 2, the problem formulation focused on 61 food classes, and the best submission received a mAP of 0.633 and mAR of 0.886. In round 3, the problem formulation focused on 273 classes, the best submission received a mAP of 0.551 and a mAR of 0.884. In round 4, the problem formulation stayed consistent with that of round 3, and the best submission received a mAP of 0.568 and a mAR of 0.767.

The mAP scores in round 2 (0.633) are higher than that of round 4 (0.568). This is not a decrease in performance, as the classification complexity significantly increased between round 2 and round 4. For the classification task in round 2, a model has to choose between 61 food categories, while in round 4, a model has to choose between 273 food categories - making the task significantly harder. The problem formulation between round 3 and round 4 was kept consistent even if we had access to a larger number of training annotations. The reason for that was influenced by the analysis of the submissions received in the first three rounds - where participants were focused on optimizing the mAP scores. There were not any significant explorations done by the participants in the direction of trading off the mAR scores to increase the mAP scores. An immediate increase in the available training annotations would have continued the same trend, as it did across the previous rounds. But in this case, participants were forced to explore the trade offs between mAR and mAP in the final focused phase of the first iteration of the Food Recognition Benchmark.

Additionally, Figure E.1 shows the distribution of the mAP scores from the submitted results across time, and across all the four rounds of the benchmark. Figure E.1 shows the distribution of the mAP scores and the mAR scores across all the submissions made in the benchmark. It is to be noted that the scores across round 1, round 2, and rounds 3 and 4 are not immediately comparable as they have different number of food categories, which is an artifact of any evolving benchmark.

|  | Number of Food Categories | Number of Training Images | Number of Training Annotations | mean Average Precision IoU >0.5 | mean Average Recall IoU >0.5 |
| --- | --- | --- | --- | --- | --- |
| Round 1 | 40 | 5545 | 7735 | 0.573 | 0.831 |
| Round 2 | 61 | 7949 | 11468 | 0.633 | 0.886 |
| Round 3 | 273 | 24119 | 39325 | 0.551 | 0.884 |
| Round 4 | 273 | 24119 | 39325 | **0.568** | 0.767 |

**Table 1.** mean Average Precision (mAP) and mean Average Recall (mAR) scores from the best submissions received across the four rounds of the Food Recognition Benchmark.

|  | Backbone | Experiment | mean Average Precision$_{IoU>0.5}$ | mean Average Recall$_{IoU>0.5}$ |
| --- | --- | --- | --- | --- |
| **Mask RCNN** | ResNet50 | Baseline | 0.473 | 0.707 |
|  |  | MultiScale Training | 0.482 | 0.732 |
|  |  | Weighted Loss | 0.479 | 0.713 |
|  |  | Train Time Augmentation | 0.487 | 0.741 |
|  |  | Combined | 0.506 | 0.809 |
|  | ResNet101 | Baseline | 0.485 | 0.706 |
|  |  | Combined | 0.523 | 0.817 |
|  | ResNeXt101 | Baseline | 0.474 | 0.710 |
|  |  | Combined | 0.535 | 0.825 |
| **Hybrid Task Cascade (HTC)** | ResNet50 | Baseline | 0.484 | 0.800 |
|  |  | Combined | 0.525 | 0.861 |
|  | ResNet101 | Baseline | 0.491 | 0.798 |
|  |  | Combined | **0.539** | 0.867 |

**Table 2.** mean Average Precision (mAP) and mean Average Recall (mAR) scores from the Mask RCNN and Hybrid Task Cascade(HTC) experiments across different experimental configurations on the Round 4 problem formulation and test set.

However, the scores in round 3 and round 4 are comparable as they focus on the same problem formulation.

***E.2. Experiment Results.*** Table 2 outlines the mAP and mAR scores from the Mask RCNN and Hybrid Task Cascade (HTC) experiments across different experimental configurations. The experiments were conducted on the Round 4 problem formulation and evaluated on the Round 4 Test set.

In the first category of experiments, we explore the Mask RCNN model with multiple backbones. For a Mask RCNN model with a ResNet50 backbone, we report results separately from experiments with the vanilla baseline model, with multi scale training, training with weighted loss, and training with augmentations. While the results in each of the individual experiments are comparable, with the baseline experiment scoring a mAP score of 0.473 and a mAR score of 0.707; in case of the combined experiment where we include all the above mentioned aspects during the training, we get the best results for Mask RCNN models with a mAP score of 0.506 and a mAR score of 0.809. The performance gain in the combined experiments is consistent across all experiment categories. In the rest of the experiment categories we report the results from just the baseline model and the combined experiments.

Across the MaskRCNN experiments, as expected, the best results are obtained with a ResNeXt101 backbone with a mAP score of 0.535 and a mAR score of 0.825; closely followed by the ResNet101 backbone with a mAP score of 0.523 and a mAR score of 0.817; and finally the ResNet50 backbone with a mAP score of 0.506 and a mAR score of 0.809.

The second category of experiments explore the Hybrid Task Cascade (HTC) models, which outperform the MaskRCNN models, with the ResNet50 backbone obtaining a mAP score of 0.525 and a mAR score of 0.861, and the ResNet101 backbone obtaining the best mAP score of 0.539 and a mAR score of 0.867.

The experimental results reported are from the individual experiments and do not include results from ensembles across the individual models. The code for reproducing the experiments reported in this manuscript are available at https://gitlab.aicrowd.com/aicrowd/research/myfoodrepo-experiments.

**F. Conclusion.** We introduce a novel instance segmentation dataset for real world images from 273 food classes. We reported here the results of the first four rounds of the AIcrowd Food Recognition Benchmark. The goal of the benchmark is to create open, stable and reproducible food recognition algorithms for broad use. The benchmark provided 24,119 publicly available images taken by MyFoodRepo users and annotated by professional annotators on the MyFoodRepo web platform. Submitted algorithms were evaluated on private images sourced from the same app / platform. The strength of the underlying dataset is that it has been collected in the context of a personalized health cohort of generally healthy subjects, and thus represents a visually unbiased sample of food images (i.e. the images have not been selected to match any visual criteria, as may be the case in datasets from websites or social networks). The benchmark attracted 1065 participants (as of 17th April, 2021) from 71 countries, who made a total of 2,603 submissions.

The results are of direct applied use as the top-performing algorithm is made available through an API. Since June 2021, the myFoodRepo app, from which the images of the benchmark have been sourced, is using this API to annotate new images. As described in the main text, these automatic annotations are then verified - and corrected if necessary - by human annotators. Because the human annotation step is the most time-consuming part of the food annotation pipeline, improved models are expected to reduce this bottleneck. Preliminary measurements on the MyFoodRepo platform indicate that switching to the top-performing model from this benchmark has resulted in a significant reduction in annotation time (data not shown).

As the MyFoodRepo app continues to be used, and is planned to be used in cohorts outside of Switzerland, future versions of the benchmark are expected to be based on an ever growing size and diversity of the dataset. In addition to classification and segmentation, we hope to be able to also address the problem of volume / weight estimation, a significant challenge in food recognition.

**G. Acknowledgement.** We are grateful for the generous support of the Seerave Foundation in enabling this work. The MyFoodRepo project and the Swiss Food & You cohort where the images were sourced received generous support from the Kristian Gerhard Jebsen Foundation and the Leenaards Foundation. We thank the MyFoodRepo users who created the data sets by agreeing to make their images publicly available. We further thank all participants of the food recognition challenge.

**H. Conflict of Interest.** Authors SPM and MS are co-founders of AIcrowd. ES and GS have been among the top performing participants and have been invited to coauthor the paper. They declare no conflict of interest. The remaining authors declare that the research was conducted in the absence of any commercial or financial relationships that could be construed as a potential conflict of interest.

1. GBD 2017 Diet Collaborators, et al., Health effects of dietary risks in 195 countries, 1990–2017: a systematic analysis for the global burden of disease study 2017. *The Lancet* **393**, 1958–1972 (2019).
2. W Willett, et al., Food in the anthropocene: the eat–lancet commission on healthy diets from sustainable food systems. *The Lancet* **393** (2019).
3. D Zeevi, et al., Personalized nutrition by prediction of glycemic responses. *Cell* **163**, 1079–1094 (2015).
4. BB Finlay, et al., Can we harness the microbiota to enhance the efficacy of cancer immunotherapy? *Nat. Rev. Immunol.* **20**, 522–528 (2020).
5. JL McQuade, CR Daniel, BA Helmink, JA Wargo, Modulating the microbiome to improve therapeutic response in cancer. *The Lancet Oncol.* **20**, e77–e91 (2019).
6. D Davar, et al., Fecal microbiota transplant overcomes resistance to anti–PD-1 therapy in melanoma patients. *Science* **371**, 595–602 (2021).
7. R Hills, et al., Gut microbiome: Profound implications for diet and disease. *Nutrients* **11**, 1613 (2019).
8. S Downer, SA Berkowitz, TS Harlan, DL Olstad, D Mozaffarian, Food is medicine: actions to integrate food and nutrition into healthcare. *BMJ*, m2482 (2020).
9. Y Matsuda, H Hoashi, K Yanai, Recognition of multiple-food images by detecting candidate regions in *Proc. of IEEE International Conference on Multimedia and Expo (ICME)*. (2012).
10. Y Kawano, K Yanai, Automatic expansion of a food image dataset leveraging existing categories with domain adaptation in *Proc. of ECCV Workshop on Transferring and Adapting Source Knowledge in Computer Vision (TASK-CV)*. (2014).
11. T Ege, K Yanai, A new large-scale food image segmentation dataset and its application to food calorie estimation based on grains of rice in *Proc. of ACMMM Workshop on Multimedia Assisted Dietary Management(MADiMa)*. (2019).
12. K Okamoto, K Yanai, UEC-FoodPIX Complete: A large-scale food image segmentation dataset in *Proc. of ICPR Workshop on Multimedia Assisted Dietary Management(MADiMa)*. (2021).
13. D Pishva, A Kawai, K Hirakawa, K Yamamori, T Shiino, Bread recognition using color distribution analysis. , 1651–1659 (2001).
14. G Ciocca, P Napoletano, R Schettini, Food recognition: A new dataset, experiments, and results. *IEEE J. Biomed. Heal. Informatics* **PP**, 1–1 (2016).
15. E Aguilar, B Remeseiro, M Bolaños, P Radeva, Grab, pay, and eat: Semantic food detection for smart restaurants. *IEEE Transactions on Multimed.* **20**, 3266–3275 (2018).
16. K He, G Gkioxari, P Dollár, R Girshick, Mask r-cnn in *2017 IEEE International Conference on Computer Vision (ICCV)*. pp. 2980–2988 (2017).
17. H Ye, Q Zou, Food recognition and dietary assessment for healthcare system at mobile device end using mask r-cnn in *Testbeds and Research Infrastructures for the Development of Networks and Communications*, eds. H Gao, K Li, X Yang, Y Yin. (Springer International Publishing, Cham), pp. 18–35 (2020).
18. CNC Freitas, FR Cordeiro, V Macario, Myfood: A food segmentation and classification system to aid nutritional monitoring in *2020 33rd SIBGRAPI Conference on Graphics, Patterns and Images (SIBGRAPI)*. pp. 234–239 (2020).
19. TY Lin, et al., Microsoft coco: Common objects in context in *European conference on computer vision*. (Springer), pp. 740–755 (2014).
20. A Chatelan, et al., Major differences in diet across three linguistic regions of switzerland: Results from the first national nutrition survey menuch. *Nutrients* **9**, 1163 (2017).
21. M Everingham, L Van Gool, CKI Williams, J Winn, A Zisserman, The PASCAL Visual Object Classes Challenge 2012 (VOC2012) Results (http://www.pascal-network.org/challenges/VOC/voc2012/workshop/index.html) (year?).
22. K Chen, et al., Hybrid task cascade for instance segmentation. , 4969–4978 (2019).
23. Z Cai, N Vasconcelos, Cascade r-cnn: high quality object detection and instance segmentation. *IEEE transactions on pattern analysis machine intelligence* (2019).
24. S Qiao, L Chen, AL Yuille, Detectors: Detecting objects with recursive feature pyramid and switchable atrous convolution. *CoRR* **abs/2006.02334** (2020).
25. K He, X Zhang, S Ren, J Sun, Deep residual learning for image recognition. *arXiv preprint arXiv:1512.03385* (2015).
26. S Ren, K He, RB Girshick, J Sun, Faster R-CNN: towards real-time object detection with region proposal networks. *CoRR* **abs/1506.01497** (2015).
27. EKVII A. Buslaev, A. Parinov, AA Kalinin, Albumentations: fast and flexible image augmentations. *ArXiv e-prints* (2018).
28. Á Casado-García, et al., CLoDSA: a tool for augmentation in classification, localization, detection, semantic segmentation and instance segmentation tasks. *BMC Bioinforma.* **20** (2019).